\documentclass[conference,compsoc]{IEEEtran}

\usepackage{amsmath}
\usepackage{filecontents}
\usepackage{microtype}
\usepackage{hyperref}
\usepackage{url}
\usepackage{booktabs}
\usepackage{amsmath}
\usepackage[most]{tcolorbox}
\usepackage{xcolor}
\usepackage[table]{xcolor} 
\usepackage{multirow}
\definecolor{lightpurple}{RGB}{240,240,255}
\usepackage{lineno,ulem}
\usepackage{enumitem}
\usepackage{array}
\newcommand{\alexAA}[1]{#1}
\newcommand{\alexA}[1]{#1}
\newcommand{\alex}[1]{#1}
\newcommand{\citep}{\cite}
\newcommand{\citet}{\cite}

\definecolor{darkblue}{rgb}{0, 0, 0.5}
\hypersetup{colorlinks=true, citecolor=darkblue, linkcolor=darkblue, urlcolor=darkblue}
\definecolor{softgreen}{RGB}{200, 230, 201}
\definecolor{lightgreen}{RGB}{144, 238, 144}
\definecolor{Gray}{gray}{0.9} 

\usepackage{tikz}
\usetikzlibrary{shapes.geometric, arrows.meta, positioning, calc}
\usepackage[capitalize,noabbrev]{cleveref}
\usepackage{enumitem}
\usepackage[utf8]{inputenc}
\usepackage{booktabs} 
\usepackage{longtable} 
\usepackage{booktabs}
\ifCLASSOPTIONcompsoc
  \usepackage[nocompress]{cite}
\else
  \usepackage{cite}
\fi

\definecolor{outcomecolor}{HTML}{E0F7FA}  
\definecolor{elicitcolor}{HTML}{FFFDE7}   
\definecolor{formalcolor}{HTML}{F1F8E9}   
\definecolor{purplecolor}{HTML}{F3E5F5}  

\begin{document}

\title{SoK: The Landscape of Memorization in LLMs: Mechanisms,\\ Measurement, and Mitigation
}
\author{Alexander Xiong$^1$,\quad Xuandong Zhao$^1$,\quad Aneesh Pappu$^2$,\quad  Dawn Song$^1$ \\
$^1$UC Berkeley
$^2$Google DeepMind \\
\small \texttt{\{alexxiong,xuandongzhao,dawnsong\}@berkeley.edu, aneeshpappu@google.com}
}

\newcommand{\fix}{\marginpar{FIX}}
\newcommand{\new}{\marginpar{NEW}}


\maketitle

\begin{abstract}
Large Language Models (LLMs) have demonstrated remarkable capabilities across a wide range of tasks, yet they also exhibit memorization of their training data. This phenomenon raises critical questions about model behavior, privacy risks, and the boundary between learning and memorization. Addressing these concerns, this paper synthesizes recent studies and investigates the landscape of memorization, the factors influencing it, and methods for its detection and mitigation. We explore key drivers, including training data duplication, training dynamics, and fine-tuning procedures that influence data memorization. In addition, we examine methodologies such as prefix-based extraction, membership inference, and adversarial prompting, assessing their effectiveness in detecting and measuring memorized content. Beyond technical analysis, we also explore the broader implications of memorization, including the legal and ethical implications. Finally, we discuss mitigation strategies, including data cleaning, differential privacy, and post-training unlearning, while highlighting open challenges in balancing the need to minimize harmful memorization with model utility. This paper provides a comprehensive overview of the current state of research on LLM memorization across technical, privacy, and performance dimensions, identifying critical directions for future work.
\end{abstract}

\section{Introduction}

Throughout the past few years, we have observed great strides in the capabilities of LLMs driven by changes in model architecture, training methodologies, and computational resources~\citep{radford2018improving,brown2020languagemodelsfewshotlearners,chowdhery2022palmscalinglanguagemodeling,naveed2023comprehensive,touvron2023llamaopenefficientfoundation,wei2023chainofthoughtpromptingelicitsreasoning}.  
These advances have significantly improved natural language understanding, generation, and reasoning abilities, enabling these models to perform increasingly complex tasks across diverse domains~\citep{alowais2023revolutionizing,khurana2023natural,minaee2025largelanguagemodelssurvey}. 
However, in addition to their strong capabilities, LLMs continue to manifest critical limitations when evaluated on criteria such as privacy and security~\citep{10.1145/3641289,yao2024survey,das2025security}. 

With the advent of data scale used to train LLMs, the risks of privacy leakages through memorization have dramatically increased, as models trained on massive datasets containing potentially sensitive information may inadvertently reproduce verbatim passages from training data when prompted with specific triggers, compromising individual privacy and confidentiality without explicit consent~\citep{dwivedi2023opinion,smith2023identifying,abdali2024securing,nasr2023scalableextractiontrainingdata}. 
Thus, data memorization emerges as a critical vulnerability in LLMs, presenting significant privacy risks, including potential exposure of Personally Identifiable Information (PII), copyright violations, and unauthorized reproduction of sensitive content~\citep{carlini2021extractingtrainingdatalarge,cremer2022cyber,clusmann2023future, karamolegkou2023copyright,song2023digital,freeman2024exploringmemorizationcopyrightviolation,hua2024limitations, morris2025much}. 

This investigation explores memorization mechanisms in LLMs, examining contributing factors, detection methodologies, measurement approaches, and mitigation techniques. Unlike previous surveys on security and privacy~\citep{siau2020artificial,fui2023generative,guo2023evaluating,hadi2023large,karabacak2023embracing,min2023recent, 
zhao2023survey,zhu2023large,10.1145/3641289,liu2024datasets,myers2024foundation,raiaan2024review,stahl2024ethics,zhao2024explainability}, 
we present a focused examination of memorization phenomena in LLMs, dissecting the fundamental components contributing to and mitigating unintended data retention~\citep{bender2021dangers, zhao2022provably, hartmann2023sok, satvaty2024undesirable}. 
Our work provides unique contributions that distinguish it from existing surveys. We significantly expand coverage of existing work presented in~\citet{ hartmann2023sok, satvaty2024undesirable}. While \alexAA{Hartmann, et al.}~\citet{hartmann2023sok} focused on definitions, implications (e.g., model performance and alignment, privacy, security and confidentiality, copyright, and auditing), detection, and prevention, this paper provides 
 an in-depth review of the existing research on LLM memorization and its outcomes by emphasizing concept differences and relationships, main research 
 challenges with key barriers and research directions at different memorization stages, 
 the key principles, requirements and practicality, and core limitations and statistical soundness of detection techniques, 
 and research challenges for detection, mitigation, and privacy and legal risk of memorization. In contrast with the survey by \citet{satvaty2024undesirable}, our analysis examines the technical underpinnings, evaluation methods, and privacy implications of memorization and provides concrete research directions. We organize open research questions systematically by memorization subtopics and provide some rationale before presenting each question. While identifying open problems is inherently subjective, we select the most pressing issues based on current research challenges, though we acknowledge that other researchers may prioritize different questions and that these priorities will likely evolve as the field advances.

\begin{figure*}[t]
\label{memorization-taxonomy}
    \centering
    \includegraphics[width=\linewidth]{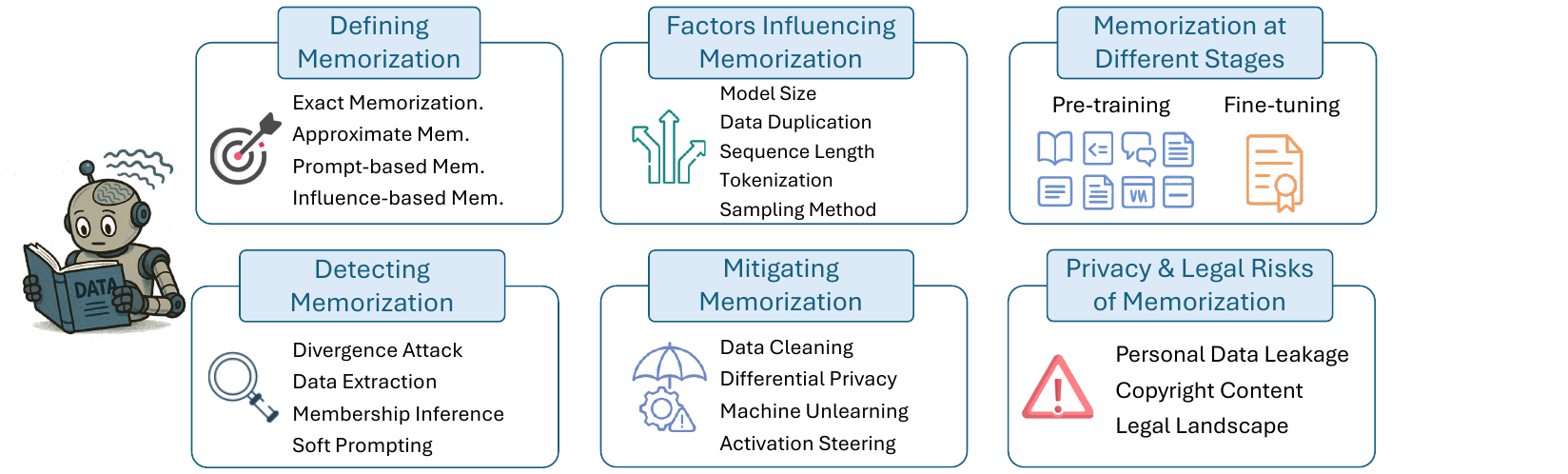}
 \caption{\footnotesize Taxonomy of understanding memorization in LLMs.}
  \label{fig:mem_fig}
\end{figure*}

Figure \ref{memorization-taxonomy} provides a visual taxonomy that maps and categorizes memorization, detection and mitigation of memorization, and corresponding privacy and legal risks. 

\section{Defining memorization}

\alexAA{
This research examines cases of unintended memorization, where LLMs reproduce sensitive, private, or verbatim training data. While memorizing facts or grammatical rules is essential for model utility, the inadvertent reproduction of private or non-public data raises significant privacy and ethical concerns. The central challenge, which this paper aims to systematize, is distinguishing between beneficial and harmful forms of memorization. This work focuses specifically on the latter, providing a comprehensive overview of the mechanisms, detection methods, and mitigation strategies for unintended data reproduction in LLMs.}

Agnostic to LLMs, memorization occurs when a model can reproduce specific training sequences when provided with appropriate contextual inputs~\citep{carlini2023quantifyingmemorizationneurallanguage, cooper2023report, schwarzschild2024rethinking}. \alex{This is the type of memorization most commonly studied in the context of generative models.}
Currently, several definitions are used to quantify memorization in LLMs, including memorization defined by string matching (exact or approximate) or by loss differences. In addition, \alexAA{Hayes, et al.}~\citet{hayes2025measuringmemorizationlanguagemodels} employs baseline comparisons to distinguish genuine memorization from coincidental sequence generation.

However, it is important to recognize that memorization is not inherently negative; LLMs also depend on the beneficial memorization of factual, linguistic, and semantic regularities that underpin generalization, coherence, and factual accuracy. Specifically, memorization contributes positively to a model's ability to store and retrieve knowledge. The critical distinction lies in what is memorized and how it is used. 

Appendix's Section 1
expands on these memorization definitions. \alex{The heterogeneity of these definitions across the literature has resulted in a fragmented landscape of detection and mitigation strategies. This lack of a unified evaluation framework is a key challenge, as existing definitions are constrained and exhibit considerable overlap, hindering the direct comparison of results across studies. To provide clarity, Table~\ref{table-def} systematizes these concepts, outlining their key differences, interrelationships, and 
contributions and effects.
}

\begin{table*}[h!]
\footnotesize
\centering
\label{table-def}
\resizebox{\linewidth}{!}{
\begin{tabular}
{p{1.8cm}|p{2.8cm}|p{3.5cm}|p{5cm}|p{5.9cm}} 
\toprule
{\bf Category} & {\bf Definition} & {\bf Defining Characteristic} & {\bf Relationship to Other Definitions} & {\bf 
Contribution \& Effect} \\
\midrule
\rowcolor{softgreen} 
{\bf Outcome-Centric} & 
{\bf Verbatim / Perfect Memorization} & 
Exact, word-for-word reproduction of a sequence from the training data, with no deviation. & 
The most stringent outcome. It is the target output for Eidetic recall and a successful instance of Extractable/Discoverable memorization. Directly opposes Approximate Memorization. & 
{\bf Contribution:} Establishes a baseline for the most direct and severe form of data leakage. \newline {\bf Effect:} Forms the legal and privacy basis for concerns about copyright infringement and PII exposure. \\
\midrule
\rowcolor{softgreen} 
{\bf Outcome-Centric} &
{\bf Approximate / Paraphrased Memorization} & 
Generation of text that is semantically equivalent to a training sequence but not a verbatim copy (e.g., rephrased). & 
A relaxation of Verbatim memorization. Its detection is a known challenge for most Elicitation-based methods, which often rely on exact string matching. & 
{\bf Contribution:} Broadens the scope of privacy risks beyond exact copies to include intellectual property and sensitive information leakage. \newline {\bf Effect:} Highlights the limitations of simple string-matching detection methods. \\
\midrule
\rowcolor{softgreen} 
{\bf Outcome-Centric} &
{\bf Eidetic Memorization} & 
The high-fidelity, verbatim recall of long, complex, or low-probability sequences. & 
A qualitative descriptor for exceptionally strong instances of Verbatim and Extractable memorization. It represents the upper bound of recall fidelity. & 
{\bf Contribution:} Characterizes a particularly strong form of memorization that implies deep embedding of data rather than surface-level overfitting. \newline {\bf Effect:} Signals a deeper, more concerning level of data retention by the model. \\
\midrule
\rowcolor{elicitcolor}
{\bf Elicitation-Centric} & 
{\bf Extractable Memorization} & 
The existence of \textit{any} constructible prompt that causes the model to generate a specific training example. & 
The broadest elicitation category. It serves as a superset for Discoverable, $k$-extractable, and (prompt-triggered) Verbatim memorization. & 
{\bf Contribution:} Defines the absolute attack surface; if data can be retrieved, it is considered a vulnerability, regardless of prompt likelihood. \newline {\bf Effect:} Models the worst-case potential for data extraction. \\
\midrule
\rowcolor{elicitcolor}
{\bf Elicitation-Centric} &
{\bf Discoverable Memorization} & 
Elicitation of a training sequence's suffix by prompting the model with its corresponding prefix. & 
A practical and constrained subset of Extractable Memorization. Stronger than general Extractable but weaker than Verbatim, as the output may not always be an exact match. & 
{\bf Contribution:} Provides a measurable and replicable method for probing memorization using known training data prefixes. \newline {\bf Effect:} Enables systematic, large-scale privacy auditing and unwanted ``auto-completion" risk analysis. \\
\midrule
\rowcolor{elicitcolor}
{\bf Elicitation-Centric} &
{\bf $k$-extractable Memorization} & 
A refinement of Discoverable where the prefix has a specified length, $k$. & 
A more formalized and measurable subtype of Discoverable and Extractable memorization. & 
{\bf Contribution:} Allows for quantifying the relationship between context length ($k$) and the likelihood of extraction. \newline {\bf Effect:} Used to analyze how prompt length affects memorization recall. \\
\midrule
\rowcolor{purplecolor} 
{\bf Probabilistic}
& 
{\bf ($n$, $p$)-Discoverable Extraction} & 
A probabilistic formalization where a sequence is considered memorized if it is generated at least once in $n$ attempts with probability $\ge p$. & 
A probabilistic extension of Discoverable and $k$-extractable that accounts for the stochastic nature of sampling-based decoding. & 
{\bf Contribution:} Moves beyond deterministic checks to a more realistic, probabilistic framework for assessing leakage in real-world, non-greedy decoding scenarios. \newline {\bf Effect:} Better quantifies data exposure risk under diverse sampling strategies. \\
\midrule
\rowcolor{outcomecolor} 
{\bf Causal} &
{\bf Counterfactual Memorization} & 
A causal definition where a model's output is attributed to memorization only if the output changes upon removing the specific data point from the training set. & 
Fundamentally different from other definitions by requiring a causal link, not just correlational output. It provides a theoretical grounding for all other forms of memorization. & 
{\bf Contribution:} Isolates ``true" memorization from ``coincidental" generation (i.e., the model would have generated it anyway). \newline {\bf Effect:} Allows researchers to causally separate memorization from learned generalization, which is crucial for unlearning methods and for legally defensible audits of data influence. \\
\midrule
\rowcolor{Gray} 
{\bf Efficiency \& Info-Theoretic} &
{\bf $\tau$-Compressible Memorization (ACR)} & 
Memorization defined by the compression ratio ($\tau$) between the prompt length and the output length. High compression indicates efficient extraction. & 
An information-theoretic approach to Extractable's objective. It provides a continuous spectrum of memorization ``strength" rather than a binary classification. & 
{\bf Contribution:} Offers a quantifiable metric for extraction efficiency. \newline {\bf Effect:} Potentially relevant for fair use legal arguments, by linking the ``effort" (prompt complexity) to the ``recall" (output length). \\
\bottomrule
\end{tabular}
}\vspace{2mm}
\caption{A Taxonomy of Memorization Definitions in Large Language Models}
\end{table*}

\section{Factors influencing memorization}

The memorization behavior of LLMs is governed by a range of factors spanning training and inference.

\textbf{Model parameter size.}
Model size strongly correlates with increased memorization in LLMs, with larger models demonstrating both greater capacity to retain training data and increased vulnerability to extraction attacks. \alexAA{Carlini, et al.}~\citet{carlini2021extractingtrainingdatalarge} first introduced the relationship between model size and memorization, observing that memorization scales log-linearly with model size. Subsequent research has consistently reinforced this finding across multiple dimensions of memorization. For instance, \alexAA{Tirumala, et al.}~\citet{tirumala2022memorizationoverfittinganalyzingtraining} observed that larger LLMs not only memorize more content but do so more rapidly during training, a phenomenon that cannot be fully explained by conventional overfitting or hyper-parameter tuning. This enhanced memorization capability directly impacts privacy concerns, as \alexAA{Li, et al.}~\cite{li2024llmpbeassessingdataprivacy} demonstrated that extraction attacks become significantly more effective against larger LLMs due to their advanced memorization capabilities. The scaling effect has been further corroborated by~\citep{ lee2022deduplicatingtrainingdatamakes, tirumala2022memorizationoverfittinganalyzingtraining,carlini2023quantifyingmemorizationneurallanguage,freeman2024exploringmemorizationcopyrightviolation, kiyomaru-etal-2024-comprehensive-analysis,li2024llmpbeassessingdataprivacy, hayes2025measuringmemorizationlanguagemodels, morris2025much}.

\textbf{Training data duplication.}
Duplicate examples in training data skew LLMs toward overrepresented content, diminishing output diversity and increasing verbatim reproduction of training material~\citep{carlini2021extractingtrainingdatalarge}. As demonstrated in~\citet{lee2022deduplicatingtrainingdatamakes}, de-duplication substantially reduces memorization; models trained on original data exhibited a tenfold decrease in memorized token generation compared to those trained on datasets where exact substrings and near-duplicates were not removed. Exploring further, \alexAA{Kandpal, et al.}~\citet{kandpal2022deduplicatingtrainingdatamitigates} identified a superlinear relationship between training data duplication and memorization in LLMs, where rarely duplicated training samples are seldom memorized. 

\alexAA{Current deduplication techniques encode a compromise between human intuition about content redundancy and the computational efficiency required at a massive scale. Consequently, studies show that memorization often persists after syntactic deduplication~\cite{lee2022deduplicatingtrainingdatamakes}, driven primarily by near-duplicates (e.g., paraphrased sentences or text with minor edits) that fall below the similarity thresholds currently in use~\cite{SHSP2024}. This suggests the need for training data attribution–based approaches that can analyze the influence of near-duplicates and identify which content segments the model internally treats as equivalent, rather than relying solely on surface-level hash or substring matching. Adopting such a model-centric perspective would redefine duplication based on the learned internal representations of the model itself, rather than on pre-defined similarity heuristics. For example, recent studies have successfully used influence functions~\cite{KohLiang2017Understanding} to quantify the contribution of individual training instances (e.g., near-duplicates) to specific memorized outputs, confirming that models can treat syntactically distinct but semantically related data as equivalent~\citep{menta2025analyzing}.
} Understanding this internal behavior is crucial, as it helps explain why even highly duplicated sequences are generated at rates significantly lower than perfect memorization would predict, suggesting that other complex dynamics, not just data frequency, govern the recall process.

\textbf{Sequence length.}
Longer sequences increase memorization in LLMs. \alexAA{Carlini, et al.}~\citet{carlini2023quantifyingmemorizationneurallanguage} found that memorization increases logarithmically with sequence length, with the verbatim reproduction probability rising by orders of magnitude as sequences extend from 50 to 950 tokens. \alex{Specifically, \alexAA{Carlini, et al.}~\citet{carlini2023quantifyingmemorizationneurallanguage} showed that the length of the prefix used as a prompt matters.} Similarly, extraction methodologies, including continuous soft prompting and dynamic soft prompting techniques, demonstrate a consistent pattern wherein the volume of extracted data grows proportionally with increasing prefix token sizes~\citep{wang-etal-2024-unlocking}.

\textbf{Tokenization.}
Models trained with larger Byte Pair Encoding (BPE) vocabularies were found to memorize significantly more sequences from their training data. \citet{kharitonov2021bpe} studied how different vocabulary sizes, produced via BPE, affect a transformer LLMs’ tendency to memorize sequences. Memorization was particularly strong for named entities, URLs, and uncommon phrases, often becoming single tokens under larger BPE settings.

\textbf{Sampling methods.}
Sampling methods significantly impact the extraction of memorized content from LLMs, with stochastic approaches consistently outperforming greedy decoding in revealing memorized training data. While \alexAA{Carlini, et al.}~\citet{carlini2021extractingtrainingdatalarge} initially established that greedy decoding can reveal memorized sequences but misses others due to its low diversity, subsequent research has demonstrated the superior effectiveness of other sampling methods. \alexAA{Yu, et al.}~\citet{yu2023bag} showed that optimizing sampling parameters like top-k, nucleus sampling, and temperature can substantially increase memorized data extraction, in some cases doubling previous baselines. This finding was reinforced by Tiwari and Suh~\citet{tiwari2025sequencelevelleakagerisktraining}, who discovered that randomized decoding nearly doubles leakage risk compared to greedy decoding, contradicting earlier assumptions. \alexAA{Hayes, et al.}~\citet{hayes2025measuringmemorizationlanguagemodels} provided a theoretical foundation for these observations by introducing a probabilistic framework that demonstrates how repeated sampling with varied decoding parameters can expose memorization hidden under greedy approaches. Given that no single decoding method minimizes leakage across all scenarios, \alexAA{Tiwari and Suh}~\citet{tiwari2025sequencelevelleakagerisktraining} emphasized the importance of assessing memorization under diverse sampling strategies.

\textbf{Data modality and content characteristics.} The influence of data duplication and other factors is not uniform across different types of data. Research into multi-modal systems shows that memorization risks vary across modalities like text, images, and audio~\citep{aghabagherloo2025impact}. For example, diffusion models have been shown to replicate images from their training set, a form of visual memorization~\citep{somepalli2023understanding}, and vision-language models can exhibit cross-modality memorization, where a textual prompt elicits a memorized image or vice-versa~\citep{wen2025quantifyingcrossmodalitymemorizationvisionlanguage}. Beyond modality, the inherent characteristics of the data play a crucial role. Unique or novel sequences, such as PII or secret keys, are more susceptible to memorization because the model cannot generalize from other examples and must store the information explicitly to minimize loss on that data point\citep{feldman2020neural}.

\textbf{Interplay of factors.} The factors influencing memorization are intertwined, creating a complex system where their effects are often synergistic and non-stochastic. This process can be imagined as a cascade with data duplication providing a strong initial signal, which is then captured by the large parameter size of models, a relationship well-described by scaling laws~\citep{hernandez2022scaling}. This is exacerbated by tokenization, which can collapse lengthy, unique sequences into single, easily memorized units, directly impacting how sequence length contributes to memorization risk. Specific prompts act as keys while varied sampling methods can either reveal the most obvious memorized content or, over repeated trials, uncover a far wider range of subtly store information~\citep{hayes2025measuringmemorizationlanguagemodels}. As further discussed in Section~\ref{sec:memorization-at-different-stages}, these dynamics can be unified through the lens of data compression, where models learn to memorize predictable sequences as an efficient strategy to minimize training loss~\citep{schwarzschild2024rethinking}. This latent memorization is shaped by the temporal training dynamics that favor recently seen data~\citep{jagielski2023measuring}. This intricate interplay of factors means that memorization often arises from a ``perfect storm" of factors, underscoring that various confounding factors may determine the strength of memorization as well as that effective mitigation requires a holistic approach rather than addressing any single factor in isolation.

While data duplication has been broadly associated with increased memorization, it remains uncertain whether this effect systematically varies across different data modalities (e.g., code, prose). Additionally, although inference hyper-parameters, such as decoding methods, are known to 
influence memorization, existing techniques do not reliably prevent the extraction of memorized content. This underscores the need for decoding strategies explicitly designed to minimize memorization. In contrast, the influence of training hyper-parameters on memorization remains poorly understood, particularly with regards to the role of overfitting through learning rate and regularization mechanisms. \alexAA{Pappu, et al.}~\citet{pappu2024measuringmemorizationrlhfcode} noted that larger learning rates can increase memorization, but within the narrow domain of Reinforcement Learning from Human
Feedback (RLHF) for code they investigate. Whether this is simply equivalent to whether the model well-fits the training data (i.e., under or overfits) is unknown. Moreover, the temporal dynamics of memorization in the presence of data duplication remain underexplored. Specifically, it is unclear how the timing of when duplicated examples are encountered during training affects memorization likelihood.

\begin{tcolorbox}[
    colback=lightpurple,
    colframe=black,
    title=\textbf{Open Questions},
    fonttitle=\bfseries,
    coltitle=white,
    colbacktitle=black,
    enhanced,
    sharp corners=south,
    boxrule=0.8pt
]
1. Beyond duplication, what intrinsic data properties govern memorization?\\
2. What is the optimal trade-off between memorization prevention vs. utility at inference?\\
3. How does model architecture scale influence the mechanisms of memorization?\\
4. How can we distinguish useful generalization from memorization in LLMs?
\end{tcolorbox}

\textbf{OQ1}. Since models memorize not only duplicated but also unique sequences, memorization risk appears linked to intrinsic data properties like low complexity or high predictability. The core research challenge is to formalize a predictive theory of this ``memorization potential," enabling proactive data curation to identify and remove high-risk content prior to training.

\textbf{OQ2}. Decoding-time defenses rely on a real-time ``leakage score" to guide text generation away from memorized training data~\citep{ippolito-etal-2023-preventing}. The open frontier is to formulate a metric that is both computationally tractable for inference and can navigate the fundamental trade-off between preventing harmful regurgitation and maintaining model utility.

\textbf{OQ3}. It remains unclear if larger models simply have more capacity for memorization or if the internal mechanisms themselves evolve with scale. Tracing ``memorization circuits"~\citep{meng2022locating} within components like FFN layers theorized to act as key-value memories~\citep{geva2022transformer} is crucial for understanding how these circuits change in smaller models and for designing future architectures that are inherently safer.

\textbf{OQ4}. The ideal test for memorization involves a counterfactual~\citep{zhang2023counterfactual}: would the model still produce this output if this exact training example were removed? Since retraining a massive model to answer this is computationally impossible, the critical research frontier is developing scalable approximations that provide a practical, operational line between useful knowledge and harmful regurgitation.

\section{Memorization at different stages} 
\label{sec:memorization-at-different-stages}
Although memorization has often been attributed to overfitting, this section explores how it is more deeply shaped by pre-training dynamics and fine-tuning strategies in LLMs.

\textbf{Pre-training dynamics} introduce systematic biases that influence which training examples are retained. Under standard non-deterministic training regimes, characterized by data shuffling, dropout, and stochastic optimization, models tend to forget examples seen early in training unless they are revisited, as shown in~\citet{kiyomaru-etal-2024-comprehensive-analysis, jagielski2023measuring,leybzon2024learning}. They hypothesize that this forgetting is not simply due to limited exposure but rather the result of parameter drift, whereby updates driven by later data overwrite earlier representations. Consequently, memorization is biased toward examples encountered in the latter stages of training.

This effect becomes even more pronounced as models approach the end of training. \alexAA{Huang, et al.}~\citet{huang-etal-2024-demystifying} showed that later-stage checkpoints are more susceptible to memorizing even rare or out-of-distribution content, likely reflecting increased model capacity and representational flexibility at those stages. Extending this perspective, \alexAA{Biderman, et al.}~\citet{biderman2023emergentpredictablememorizationlarge} demonstrated that memorization follows predictable scaling laws. As model size and training duration increase, specific sequences predictably transition from unmemorized to memorized. This behavior reinforces that memorization should not be viewed as a simple artifact of overfitting, but rather as an emergent property of scale that is heavily exacerbated by, and often inseparable from, it.

\alexAA{Crucially, the field's inconsistent use of memorization metrics creates a methodological schism that threatens the validity of many empirical findings. Verbatim completion and Membership Inference Attacks (MIAs) do not merely offer different views; they can deliver contradictory verdicts on whether a model remembers or forgets. The work of Tirumala, et al.\citet{tirumala2022memorizationoverfittinganalyzingtraining} exposes this paradox: a model can be shown to ``forget" a sequence through the lens of MIAs while simultaneously reproducing it flawlessly upon prompting. The culprit is data duplication, which creates a weak, distributed membership signal that evades MIA detection while solidifying the sequence as a trivial, high-probability output. This methodological divergence is so severe that any claim about memorization dynamics, including the role of data duplication\citep{jagielski2024} or the rate of forgetting, is effectively ill-defined without a precise specification of the metric. This necessitates a critical re-evaluation of prior results and calls for immediate community-wide efforts toward standardized memorization evaluation protocols.}

\textbf{Supervised fine-tuning} influences memorization, shaping both the extent and nature of information retention and exposure. \alexAA{Mireshghallah, et al.}~\citet{mireshghallah-etal-2022-empirical} provided a comparative analysis of fine-tuning approaches, demonstrating that head-only fine-tuning presents the highest risk of memorization, likely due to overfitting. In contrast, adapter-based fine-tuning, when constraining parameter updates~\citep{zeng2024exploringmemorizationfinetunedlanguage}, reduces memorization by using parameter-efficient methods, indicating that fine-tuning can be effective for limiting privacy risks. 
They link differences in memorization from models fine-tuned for tasks such as summarization vs. question answering to the attention dynamics of each task, with narrower attention patterns correlating with higher memorization. From an adversarial perspective,  \alexAA{Nasr et al.}~\citet{nasr2023scalableextractiontrainingdata} and \alexAA{Chen et al.}~\citet{chen2024janusinterfacefinetuninglarge} demonstrated that fine-tuning can also be leveraged to extract memorized pretraining data. Using the Janus Interface, they showed that targeted fine-tuning amplifies a model's capacity to leak sensitive information, thus framing fine-tuning as a mechanism that can reactivate latent memorization vulnerabilities. Together, these studies underscore that fine-tuning critically shapes memorization patterns.

\textbf{Reinforcement learning as post-training.} Post-training paradigms such as 
RLHF~\citep{ouyang2022training}, Reinforcement Learning with Verifiable Rewards (RLVR)~\citep{lambert2024t}, and Reinforcement Learning from Internal Feedback (RLIF)~\citep{zhao2025learning} leverage reinforcement learning to train LLMs from various feedback sources. Limited research exists on memorization propagation through reinforcement learning stages. In the context of code generation, Pappu, et al.~\citet{pappu2024measuringmemorizationrlhfcode} found that data memorized during fine-tuning persists with high frequency in post-RLHF models, while finding minimal evidence that reward model data or reinforcement learning data becomes memorized. Critical questions remain regarding the predictability of memorization persistence between fine-tuning and reinforcement learning based on data attributes.

\textbf{Distillation} is another common technique used in modern ML pipelines \citep{hinton2015distilling, zhao-etal-2022-compressing}, where a small `student' model is trained on data produced by a larger `teacher' model (typically logits). \alexAA{Chaudhari, et. al.}~\citet{chaudhari2025cascadingadversarialbiasinjection} showed that bias injected adversarially into teacher models can be amplified in student models via distillation. This naturally suggests that memorization can propagate between teacher and student models, but this has not been formally analyzed.

Table~\ref{table-stage-revised} aims to frame memorization not as a single failure, but as a dynamic risk that transforms across the LLM lifecycle. The challenges are structured to show this evolution: the risk is first introduced during pre-training due to data integrity issues like semantic duplication; reshaped into concentrated `memorization circuits' during fine-tuning; incentivized by the reward model during RLHF; and finally, propagated as a form of inheritance through distillation. This strategic breakdown clarifies that different stages present unique vulnerabilities, demanding tailored research directions, from scalable data sanitization at the start to auditable reward models at the end. This specification of research challenges intends to reveal that effective mitigation cannot be a single, post-hoc fix but requires a comprehensive defense with targeted interventions at every phase of a model's creation.

\begin{table*}[h!]
\footnotesize
\centering
\resizebox{\linewidth}{!}{
\begin{tabular}{  p{1.5cm}|p{3.8cm}|p{10.4cm} } 
\toprule
{\bf Development stages} & {\bf 
Research Challenges} & {\bf Key Barriers \& Research Directions}  \\
\midrule
Pre-training
&
How do mechanistic properties of training give rise to predictable memorization patterns?
&
{\bf Barrier:} The sheer scale of web data makes exhaustive analysis or perfect conceptual deduplication computationally infeasible. Current similarity metrics often fail to detect `invisible' semantic duplicates, which remain a primary driver of memorization~\cite{lee2022deduplicatingtrainingdatamakes, SHSP2024}.
\newline
{\bf Direction:} Shift from computationally-prohibitive post-hoc attribution to proactive, data-centric risk modeling. Focus on developing lightweight, pre-training metrics to identify and flag inherently ``memorizable" content. (e.g., based on information entropy, Kolmogorov complexity estimates, or structural patterns like PII)
\\
\midrule
 Supervised Fine-tuning (SFT)
&
How does fine-tuning on a small, narrow distribution of data alter the model's pre-trained representations to either amplify latent memorization or introduce catastrophic memorization?
&
{\bf Barrier:} Parameter-Efficient Fine-Tuning (PEFT) methods like Low-Rank Adaptation (LoRA)~\cite{hu2022lora} concentrate updates into a small subset of weights, potentially creating specialized `memorization circuits' that are hard to disentangle from circuits responsible for task-specific capabilities~\cite{hou2025impact}.
\newline
{\bf Direction:} Develop methods for provable and efficient machine unlearning tailored to the SFT context~\cite{nguyen2025survey}. Research novel regularization techniques for PEFT that penalize the verbatim memorization of SFT examples without causing catastrophic forgetting of pre-trained knowledge.
\\
\midrule
 RLHF/Post-training
&
Can the RLHF process be modeled as an adversarial game where the policy model is incentivized to find and exploit memorized knowledge from the base model to maximize reward?
&
{\bf Barrier:} The reward model acts as a black box, and its own memorized biases can implicitly encourage the policy model to reproduce undesired content~\cite{nasr2023scalableextractiontrainingdata}. The optimization process may discover that regurgitating certain memorized sequences is an effective strategy for maximizing reward~\cite{pappu2024measuringmemorizationrlhfcode}.
\newline
{\bf Direction:} Reframe alignment as an adversarial game to explicitly disincentivize memorization. Instead of simple preference optimization, co-develop a ``regurgitation-seeking" model trained to elicit memorized content from the policy model. The primary model is then rewarded for helpfulness and resisting the adversarial agent, forcing it to learn a more robust, generalized representation of knowledge.
\\
\midrule
Distillation
&
Under what conditions does knowledge distillation act as a regularizer that filters out stochastic, instance-specific memorization from the teacher, versus an amplifier that entrenches high-confidence memorized patterns into the student?
&
{\bf Barrier:} The student model is optimized to match the teacher's output distribution~\cite{hinton2015distilling}. If the teacher assigns extremely high probability to a memorized sequence, the Kullback-Leibler (KL) divergence objective directly incentivizes the student to replicate this overconfidence, facilitating the transfer of memorized content~\cite{singh2025teacher, chaudhari2025cascadingadversarialbiasinjection}.
\newline
{\bf Direction:} Research distillation objectives that go beyond matching raw logits. This includes distilling from calibrated or temperature-scaled teacher distributions to penalize overconfidence, or augmenting the loss with a regularizing term that minimizes the student's divergence from a `safe' public dataset.
\\
\bottomrule
\end{tabular}
}\vspace{3mm}
\caption{\alex{A Strategic Framework of Memorization Challenges Across Development Stages.}}
\label{table-stage-revised}
\end{table*}

Characterizing memorization across training phases requires understanding the mechanisms by which information is encoded, retained, or discarded. Although memorization tends to intensify in later training stages, the degree to which it can be attenuated or reversed remains an open question. With the increasing prevalence of fine-tuned models, it is also unclear how established memorization scaling laws extend to domain-adapted settings. Moreover, the relative contributions of training dynamics versus fine-tuning objectives to overall memorization behavior are not yet well understood.

\begin{tcolorbox}[
    colback=lightpurple,
    colframe=black,
    title=\textbf{Open Questions},
    fonttitle=\bfseries,
    coltitle=white,
    colbacktitle=black,
    enhanced,
    sharp corners=south,
    boxrule=0.8pt
]
The Lifecycle of Risk:\\
1. How do training dynamics govern the forgetting and stabilization of memorized information? \\
2. How do memorization scaling laws differ between pre-training and fine-tuning? \\
3. How can we causally attribute a memorized output to a specific training stage? \\
4. What mechanisms determine whether a pre-trained memory is reinforced or forgotten during fine-tuning? \\
5. How can knowledge be distilled from a teacher to a student without transferring memorized data?
\end{tcolorbox}

\textbf{OQ1.}
The path-dependent nature of SGD dynamics means that information learned early in training can be overwritten or ``forgotten" later~\citep{jagielski2023measuring, leybzon2024learning}. The core technical challenge is to model and predict this process of catastrophic forgetting as it applies to memorized sequences. A principled understanding would enable the design of `privacy-aware' training curricula that strategically leverage natural forgetting to unlearn sensitive data, or conversely, to stabilize essential knowledge against parameter drift.

\textbf{OQ2.}
A key technical question is whether the predictable neural scaling laws for memorization from pre-training~\citep{biderman2023emergentpredictablememorizationlarge} hold in the data-poor, overparameterized SFT regime where the model is a strong inductive prior. Investigating if the established log-linear relationships persist or if new, more aggressive scaling dynamics emerge is critical.

\textbf{OQ3.}
When a model regurgitates data, attributing the memory to either the pre-training or fine-tuning stage is a fundamental challenge. The core technical barrier is the highly non-linear dependency of the fine-tuned parameters on the initial pre-trained state, which confounds standard attribution methods like influence functions~\citep{KohLiang2017Understanding}. The research frontier is to develop novel techniques, such as multi-stage causal interventions, that can trace the provenance of a specific memory by approximating the counterfactual impact of an example's presence in either training set.

\textbf{OQ4.}
The core technical problem is to predict whether a given memory from pre-training will persist, be reinforced, or be overwritten during fine-tuning. A leading hypothesis is that persistence is governed by the geometric alignment between the fine-tuning loss gradient and the parameter subspace encoding the memory; alignment may reinforce it, while misalignment may cause it to be overwritten or become latent~\citep{nasr2023scalableextractiontrainingdata}. Validating this requires advances in mechanistic interpretability to first identify these `knowledge subspaces' and then analyze their interaction with the fine-tuning gradient field.

\textbf{OQ5.}
The standard knowledge distillation objective, which minimizes the KL divergence between student and teacher output distributions~\citep{hinton2015distilling}, poses a privacy risk. It implicitly trains the student to mimic the teacher's overconfident, low-entropy predictions characteristic of memorized training examples~\citep{singh2025teacher}. The central research goal is to design alternative distillation objectives that can successfully transfer generalized knowledge while simultaneously filtering out these memorization artifacts.

\section{Detecting memorization}

\begin{table*}[h!]
\footnotesize
\centering
\resizebox{\linewidth}{!}{
\begin{tabular}{ >{\raggedright\arraybackslash}p{3.0cm} | >{\raggedright\arraybackslash}p{3cm} | >{\raggedright\arraybackslash}p{4cm} | >{\raggedright\arraybackslash}p{6.5cm} } 
\toprule
{\bf Technique} & {\bf Key Principle} & {\bf Requirements \& Practicality} & {\bf Core Limitation \& Statistical Soundness}  \\
\midrule\midrule

\multicolumn{4}{l}{\textbf{Category: Extraction-Based} (Goal: Elicit Data)} \\ \midrule

\textbf{Divergence Attack}~\citep{nasr2023scalableextractiontrainingdata}
&
Coaxes an aligned model to revert to its pre-aligned, base model behavior.
&
\begin{itemize}[leftmargin=*, nosep]\vspace{-2mm}
    \item \textbf{Access:} Black-box
    \item \textbf{Practicality:} High (cheap prompt engineering)
\end{itemize}
&
Targets transient alignment artifacts rather than core memorization phenomena. Lacks durability and is susceptible to mitigation via model updates.
\\ \addlinespace

\textbf{Prefix-based Data Extraction}~\citep{carlini2021extractingtrainingdatalarge}
&
Queries model with a known training data prefix to elicit verbatim completion.
&
\begin{itemize}[leftmargin=*, nosep]\vspace{-2mm}
    \item \textbf{Access:} Requires training data
    \item \textbf{Practicality:} High (for auditing) / Intractable (for attack)
\end{itemize}
&
Inapplicable to black-box threat models due to its reliance on training data access. Scope is restricted to verbatim prefix completion, offering low recall for other memorization types.
\\ \midrule

\multicolumn{4}{l}{\textbf{Category: Classification-Based} (Goal: Classify Data)} \\ \midrule

\textbf{Membership Inference Attack (MIA)}~\citep{shokri2017membership,fu2024membership,huang2025df,song2025mias}
&
Classifies data as ``member" or ``non-member" based on model outputs (e.g., loss).
&
\begin{itemize}[leftmargin=*, nosep]\vspace{-2mm}
    \item \textbf{Access:} Black-box
    \item \textbf{Practicality:} Low per-instance cost, but high setup cost for calibration models.
\end{itemize}
&
Lacks statistical validity for per-instance claims due to the intractability of constructing a true null distribution. Individual classifications are uncalibrated and unsuitable as definitive evidence.
\\ \midrule

\multicolumn{4}{l}{\textbf{Category: Learned-Prompting} (Goal: Discover Prompt)} \\ \midrule

\textbf{Soft Prompting (Amplification)}~\citep{ozdayi2023controllingextractionmemorizeddata,Schulhoff2025, kim2023propile}
&
Uses gradient descent to find an optimal continuous prompt that maximizes extraction.
&
\begin{itemize}[leftmargin=*, nosep]\vspace{-2mm}
    \item \textbf{Access:} Full White-Box
    \item \textbf{Practicality:} Extremely Low (prohibitively expensive)
\end{itemize}
&
Assumes an impractical threat model. Dependency on full white-box access and prohibitive computational cost limits its applicability to research contexts, rendering it irrelevant as a practical attack vector.
\\

\bottomrule
\end{tabular}
}
\vspace{3mm}
\caption{A Comparative Analysis of Memorization Detection Techniques.}
\label{table-detection-techniques}
\end{table*}

A wide range of techniques has been developed to detect memorization in LLMs. This section categorizes and examines these methods in depth to offer a comprehensive understanding of how memorization is identified. 

\textbf{Divergence attack.}
\alexAA{Nasr, et al.}~\citet{nasr2023scalableextractiontrainingdata} proposed a divergence attack, a prompt-based extraction method that circumvents alignment defenses in instruction-tuned LLMs, coaxing the model toward reverting to pre-aligned behavior. This divergence significantly increases the likelihood of emitting memorized training data, achieving up to 150$\times$ more verbatim sequences compared to typical user queries. 
\alexAA{Nasr, et al.}~\citet{nasr2023scalableextractiontrainingdata} hypothesized that this vulnerability arises because prompts induce decoding analogous to an end-of-text token during pretraining, a context in which LLMs are known to favor high-likelihood and often memorized continuations. The attack exposes persistent memorization that may have remained latent.

\textbf{Prefix-based data extraction attack} works by querying a model with 
\alexAA{a strategic input and observing whether the model completes the rest of the sequence verbatim, or approximately, depending on the specific definition used.} 
\alexAA{The earliest demonstration of this phenomenon was by Carlini, et al.}~ \citet{carlini2021extractingtrainingdatalarge},
\alexAA{who successfully recovered hundreds of memorized examples from GPT-2. Critically, this initial work did not have access to the model's training set; instead, it used diverse, strategic prompts, such as the beginning-of-sequence token or internet data not believed to be in pretraining, to induce the model to diverge and generate training content. This demonstrated that memorization existed, but the prompt methodology was less direct than modern prefix attacks.} 

\alexAA{The systematic methodology of prompting with the initial segment of a known training sequence was pioneered by Lee et al.~\citet{lee2022deduplicatingtrainingdatamakes} in the context of showing that deduplication reduced memorization, thereby laying the groundwork for the modern prefix attack definition. Building on this result,} \alexAA{Carlini, et al.}~\citet{carlini2023quantifyingmemorizationneurallanguage} showed that longer prefixes increased the likelihood of verbatim completions by reducing ambiguity and aligning the model more strongly with memorized content. 
\alexAA{Subsequently,} \alexAA{Li, et al.}~\citet{li2024llmpbeassessingdataprivacy} introduced the LLM-PBE toolkit, a benchmark for evaluating privacy risks via prefix-based extraction. \alexAA{Li, et al.}'s findings~\citet{li2024llmpbeassessingdataprivacy} reinforced that structured prefixes, such as email headers or document beginnings, are potent at eliciting memorized sequences.

An adversarial variant of this attack was explored by~\citet{kassem2024alpaca}, who used LLMs to generate candidate prefixes likely to elicit private data from a target model. \alexAA{Kassem, et al.}~\citet{kassem2024alpaca} observed that instruction-tuned models can leak pretraining data even when the prompts diverge from the original training distribution, suggesting that memorization in LLMs may be more pervasive than what is revealed by prefix attacks. An extension of prefix attacks is soft prompting, which is explored in further detail below.


\textbf{Membership inference attack (MIA)}~\citep{shokri2017membership} has emerged as a key diagnostic tool for detecting memorization in LLMs, offering insight into whether specific data points may have been memorized during training.
To circumvent the computational overhead of training shadow models, recent work has focused on black-box strategies that exploit the model’s output probabilities or perplexity for input sequences~\citep{duan2024membershipinferenceattackswork}. 

A number of MIA techniques rely on loss- or likelihood-based metrics. \citet{yeom2018privacy} used the raw loss on a target input, assuming seen examples have lower loss. To mitigate confounding effects from input difficulty, reference-based calibration~\citep{carlini2021extractingtrainingdatalarge} subtracts the loss from a separate reference model, aiming to isolate differences due to training exposure. Another method used in~\citet{carlini2021extractingtrainingdatalarge} is zlib entropy, which normalizes loss by the zlib-compressed size of the input sequence, using compression as a proxy for complexity.
Moving beyond single-point estimates, the neighborhood attack~\citep{mattern2023membership} examines local changes in the loss landscape. By generating nearby perturbations of the target input and comparing their average loss, it identifies instances where the target appears memorized. 
Similarly, the min-$k\%$ prob method~\citep{shi2023detecting} focuses on the subset of tokens with the highest loss, under the assumption that even the most uncertain tokens in a memorized sequence will still be predicted with confidence.

Yet, despite their utility in highlighting patterns of memorization, MIAs fall short when used as per-instance indicators of training data inclusion. As \alexAA{Duan, et al.}~\citet{duan2024membershipinferenceattackswork} and \alexAA{Zhang, et al.}~\citet{zhang2025membershipinferenceattacksprove}  argued, MIAs lack a well-calibrated null model since one cannot feasibly train an identical model without the target input. This makes it impossible to meaningfully estimate false positive rates, undermining the statistical soundness of individual predictions.
This limitation is underscored by the critique in~\citet{maini2024reassessingemnlp2024s}, which challenges the claims made in~\citet{zhang2024pretraining}. The latter introduces the PatentMIA benchmark and a divergence-based calibration method for MIAs, reporting improved reliability in detecting whether a sequence appeared in pretraining. However, \alexAA{Maini, et al.}~\citet{maini2024reassessingemnlp2024s} showed these gains are likely artifacts of distributional shifts introduced by the benchmark itself, which MIAs can exploit without truly resolving the underlying calibration problem.
In light of these concerns, \alexAA{Zhang, et al.}~\citet{zhang2025membershipinferenceattacksprove} recommended that MIAs be re-framed as tools for aggregate-level privacy auditing rather than as evidence of individual training data exposure. For stronger guarantees, they advocate for alternatives, such as data extraction attacks or canary-based MIAs, which offer more direct insights into memorization and leakage.

\textbf{Soft prompting} techniques have emerged as powerful tools, enabling both the extraction and suppression of memorized content. By leveraging learned embeddings pre-pended to model inputs, continuous soft prompts can explicitly influence what a model reveals from its training data. For instance, \alexAA{Ozdayi, et al.}~\citet{ozdayi2023controllingextractionmemorizeddata} demonstrated that fixed-length continuous prompts could be trained to either amplify or diminish memorization, as measured by the extraction rates of secret sequences. Their results showed that attack prompts increased memorization leakage by up to 9.3\%, while suppress prompts decreased extraction by up to 97.7\%. Building on~\citet{ozdayi2023controllingextractionmemorizeddata}, \alexAA{Kim, et al.}~\citet{kim2023propile} introduced ProPILE, a privacy auditing framework that uses soft prompt tuning in a white-box setting to extract PII memorized by LLMs. Their learned prompts were transferable across models, suggesting consistent memorization patterns that can be systematically exploited.

Dynamic soft prompting techniques extend the static nature of continuous prompts by conditioning on the input context. \alexAA{Wang, et al.}~\cite{wang-etal-2024-unlocking} proposed prefix-conditioned prompting, where a prompt generator produces tailored soft prompts based on the input prefix. This dynamic approach allows LLMs to adapt to subtle input variations and surface context-dependent memorized completions. \alexAA{Wang, et al.}~\cite{wang-etal-2024-unlocking} revealed that dynamic prompts significantly improved the discoverable memorization rate, surpassing both static prompt and no-prompt baselines. These results underscore the role of context-sensitive prompting in uncovering latent memorized content and highlight the broader utility of soft prompting in analyzing memorization in LLMs.

\alexA{Table~\ref{table-detection-techniques} further summarizes 
the key principles, requirements and practicality, and core limitations and statistical soundness of the above detection techniques.
We observe that the divergence attack and prefix-based data extraction are both extraction attacks, as their primary goal is to compel the model to generate verbatim training data as output. In stark contrast, MIAs are score-based classifiers; they do not generate data but rather classify a given input as a training member based on model outputs like loss or perplexity. Finally, soft prompting techniques represent a distinct category of learned-prompting methods, which leverages optimization to find continuous vectors that are maximally effective at triggering memorization, separating them from attacks that use discrete, human-written prompts. These three categories (extraction, classification, and learned-prompting) serve as representative pillars in the broader landscape of detection and measurement methods.}

In Appendix's Section 2.1, 
Table~\ref{table-detect-challenges} organizes the primary research challenges in memorization detection into three distinct areas: definitional, methodological, and practical. The framework begins with the \textbf{Defining the Detection Target}, which addresses the ambiguity in what constitutes memorization (e.g., verbatim vs. paraphrased). This conceptual uncertainty leads to significant hurdles in \textbf{Methodological Rigor \& Reliability}, where the lack of standardized benchmarks and calibrated thresholds prevents consistent and comparable evaluation of detection techniques. These upstream definitional and methodological issues directly impact the practical \textbf{Scope \& Viability} of these methods, resulting in prohibitive computational costs and low precision that make large-scale, real-world auditing infeasible.
%

\textbf{Reasoning} \alex{itself is not a detection or measurement technique, but it is used to evaluate whether an LLM is genuinely reasoning or merely regurgitating memorized patterns.} Much of recent literature has been focused on training Large Reasoning Models (LRMs), models trained to reason before producing a final answer, as opposed to immediately producing an answer in their first token. To date, most memorization literature has been focused on quantifying memorization of specific datapoints by way of measuring whether the model can reproduce the training example. As the field moves towards reasoning models, a definition of what it means for a model to memorize a reasoning pattern may be useful for guiding reasoning research and ensuring reasoning models generalize. \alexAA{Huang, et al.}~\citet{huang2025mathperturbbenchmarkingllmsmath} takes a step towards this by measuring whether models trained to reason on math are able to solve slightly perturbed versions of problems in their training set. The authors find that models fail to solve problems that constitute `hard' perturbations, where the problem looks superficially similar to a problem in the training set but requires a different reasoning strategy. Creating benchmarks to measure memorization of such superficial reasoning patterns remains an open and useful direction for LRM research.

From the memorization detection strategies explored above, except for the divergence attack, adversaries either imbue or infer/know the training data intended for extraction. Therefore, it is critical to develop methods to reliably detect memorized sequences in language models without requiring access to the original training data. Furthermore, these detection methods all examine direct memorization from training data. With more powerful LLMs, the boundary between memorization and generalization is fuzzier, propelling the need to understand if memorization can be detected from patterns using reasoning.

\begin{tcolorbox}[
    colback=lightpurple,
    colframe=black,
    title=\textbf{Open Questions},
    fonttitle=\bfseries,
    coltitle=white,
    colbacktitle=black,
    enhanced,
    sharp corners=south,
    boxrule=0.8pt
]
1. How can we reliably detect memorization without training data access? \\
2. How can we attribute a model's output to in-context information versus parametric memory? \\
3. How can we detect semantic memorization? \\
4. How can we distinguish memorized reasoning shortcuts from generalizable problem-solving skills?
\end{tcolorbox}

\textbf{OQ1.}
One open challenge is building ``zero-knowledge" detectors that identify memorization from intrinsic statistical artifacts (low perplexity / high confidence), without relying on differential comparisons to training data. This requires framing detection as an out-of-distribution problem, where the goal is to distinguish plausible linguistic samples from low-complexity artifacts indicative of direct replication~\citep{schwarzschild2024rethinking}.

\textbf{OQ2.}
Distinguishing parametric recall from in-context inference is a fundamental problem of causal attribution, especially in systems like RAG where the information source is ambiguous~\citep{bai2024special}. An option is to apply mechanistic interpretability tools, such as causal tracing or activation patching~\citep{meng2022locating}, to experimentally isolate the contribution of the context window versus the model's parameters.

\textbf{OQ3.}
Current detection methods are predominantly syntactic and fail to identify the memorization of implicit semantic relationships, a key blind spot~\citep{staab2024memorizationviolatingprivacyinference}. Addressing this requires adapting knowledge probing techniques~\citep{petroni2019language} to systematically test for a suspected relational fact using a diverse set of novel, semantically equivalent queries.

\textbf{OQ4.}
A key challenge in evaluating LLM reasoning is to determine whether a model has learned a generalizable algorithm or has merely memorized a superficial ``reasoning template"\citep{xie2024memorization}. This requires specialized benchmarks designed to test for robustness. The most promising approach is to create rigorous perturbation benchmarks that systematically vary a problem's surface features (e.g., numbers, variable names) while preserving its underlying logical structure. A model's failure on such perturbed examples reveals a reliance on brittle, pattern-matching behavior rather than true procedural understanding\citep{huang2025mathperturbbenchmarkingllmsmath}.

\section{Mitigating memorization}

Mitigation techniques for memorization span various approaches, from data handling to model training and auditing. This section outlines the main categories, explores key strategies within each, and highlights their differences in effectiveness and trade-offs. 
%

\subsection{Training-Time Interventions}

\textbf{Data cleaning} plays a crucial role in mitigating memorization by limiting an LLM's exposure to overrepresented sequences.
As demonstrated by \citet{carlini2021extractingtrainingdatalarge, lee2022deduplicatingtrainingdatamakes, kandpal2022deduplicatingtrainingdatamitigates}, de-duplication minimizes the over-representation of repeated sequences. Removing these duplicates serves as a form of implicit regularization, reducing the likelihood that the LLM overfits to this redundant data.

From an optimization perspective, LLMs can minimize loss on rare training examples by memorizing. This makes PII-scrubbing effective: by removing these sequences during training,  memorization is minimized~\citep{carlini2021extractingtrainingdatalarge, jagielski2023measuring}.

\alexA{The necessity of proactive data sanitization is underscored by findings from Li, et al.\citet{li2024llmpbeassessingdataprivacy}, who demonstrated that even state-of-the-art models can leak a significant portion of held-out PII when subjected to adversarial ``true-prefix" prompts. Effective mitigation relies on systematic PII scrubbing, a process guided by frameworks such as the taxonomy developed by Lukas, et al.\citet{lukas2023analyzing}, which categorizes PII types (e.g., CARDINAL, DATE, PERSON) to enable both rule-based and machine-learning-based removal of sensitive tokens. The application of these scrubbing guidelines was shown to dramatically reduce such leakage, confirming the efficacy of targeted data cleaning as a primary defense mechanism. This principle of targeted mitigation extends beyond PII to other forms of protected content, such as copyrighted material. For instance, Vyas, et al.~\citet{vyas2023provable} introduced a method for training models to mitigate near-exact copying of copyrighted excerpts, providing provable certificates by extending differential privacy concepts to a bounded memorization budget.
}

\textbf{Differential privacy (DP)}~\citep{brown2022does}
provides robust protection against membership inference and extraction attacks, even in the presence of adversaries with auxiliary knowledge~\citep{dwork2006differential, dwork2014algorithmic, shokri2017membership, carlini2021extractingtrainingdatalarge}.
Differentially Private Stochastic Gradient Descent (DP-SGD)~\citep{abadi2016deep} guarantees privacy by computing per-example gradients, clipping them to a fixed norm to bound individual influence, and injecting calibrated Gaussian noise into the aggregated updates. A privacy accountant tracks cumulative privacy loss to enforce the global ($\epsilon$, $\delta$) budget.

\alexAA{Li, et al.}~\citet{li2021large} demonstrated that LLMs can act as strong DP learners. Their evaluation reveals that pretrained LLMs are highly resilient to the noise introduced by DP-SGD, particularly when fine-tuned on top of general-purpose representations.
DP-trained models approach the performance of non-private baselines, while offering provable protection against memorization. Empirical validation through canary insertion and MIAs confirm the effectiveness of DP in preventing leakage of sensitive sequences. To further reduce memorization, \alexAA{Zhao, et al.}~\citet{zhao2022provably} introduced Confidentially Redacted Training (CRT), a method that combines de-duplication and redaction with DP-SGD to train LLMs while avoiding the retention of sensitive content. CRT builds on DP concepts to introduce randomized training interventions that prevent unintended memorization. The authors
show that CRT, when combined with DP-SGD, provides strong privacy protections without compromising model performance, as indicated by competitive perplexity scores. \alexAA{However, integrating DP with pretraining memorization is highly complex at scale. As the provable guarantees of DP rely on bounding the ``group size," the total number of times an individual's data appears, the vast, web-scale nature of pretraining corpora, coupled with the scale duplication and near-duplicates, make reliably defining this group size extremely challenging. Consequently, the interpretation and strength of DP guarantees against pretraining-time memorization often remain unclear due to the underlying data redundancy.}

Furthermore, the use of DP in LLMs often presents trade-offs \alexAA{among} privacy, utility, and computational efficiency. Strict privacy budgets often lead to performance drops, and significant computational overhead makes it challenging to scale to LLMs. To address these challenges, recent work has examined PEFT strategies for training under DP. PEFT approaches, such as LoRA \citep{hu2022lora}, allow fine-tuning of models with addition of few parameters relative to base models via low-rank decompositions of additional trainable parameters. It is hypothesized that fine-tuning PEFT models with DP mitigates losses of utility due to requiring less noise for privatization due to training with fewer parameters \citep{liu2024differentially, yu2022differentiallyprivatefinetuninglanguage, yu2021largescaleprivatelearning}. \alexAA{Ma, et al.}~\citet{ma2024efficient} explored the integration of adapter-based fine-tuning methods into the DP framework. Their findings reveal that DP PEFT reduces memorization and achieves comparable or superior downstream task performance relative to full-model DP-SGD, under strict privacy budgets. \alexAA{Ma, et al.}~\citet{ma2024efficient} hypothesized that the limited parameter updates in PEFT may concentrate DP noise in a narrow subset of the model, potentially weakening privacy protection.
While most DP techniques for LLMs target record-level privacy, many applications require user-level guarantees. \alexAA{Chua, et al.}~\citet{chua2024mind} addressed this gap by proposing several techniques to enforce DP at user-level. 
To evaluate effectiveness, \alexAA{Chua, et al.}~\citet{chua2024mind} used canary insertion attacks, embedding unique sequences into user data to test for memorization. Results show that user-level DP reduces memorization, with much lower canary extraction rates than record-level DP.

DP in LLMs faces an efficiency-utility trade-off: strict privacy budgets degrade performance, while scaling DP methods requires significant computational overhead. Thus, the adoption of DP for mitigating memorization in LLMs remains limited.

\subsection{Post-Training-Time Interventions}
\textbf{Machine unlearning} aims to remove the influence of certain training examples so that the model’s behavior is as if those examples were never seen~\citep{cooper2024machine, liu2024unlearning}. \alexAA{Yao, et al.}~~\citet{yao2024machine} evaluates various model-editing unlearning strategies (e.g., gradient ascent~\citep{golatkar2020eternal, jia2023model, jang2022knowledge}, negative re-labeling~\citep{golatkar2020eternal}, adversarial sampling~\citep{cha2024learning}) demonstrating that approximate unlearning methods can be over $10^5\times$ more computationally efficient than retraining a model from scratch. However, unlike DP, there is no formal guarantee, thereby leaving a risk that memorization persists.

\textbf{ParaPO}, introduced by \citet{chen2025parapoaligninglanguagemodels}, is a post-training strategy to decrease memorization of data memorized in the pretraining corpus. ParaPO decreases unintended memorization by first assessing which data from the pretraining corpus has been memorized (by searching for sequences that, when a prefix is used to prompt an LM, an LM decodes a near exact match to the suffix of the sequence), and then creating preference pairs by using a separate LLM to summarize the memorized datapoint. The model is then post-trained via Direct
Preference Optimization (DPO) \citep{rafailov2024directpreferenceoptimizationlanguage} on pairs of (memorized sequence, summarization), with the summarization marked as the preferred response. \alexAA{Chen, et al.}~\citet{chen2025parapoaligninglanguagemodels} show that this approach decreases unintended memorization while preserving verbatim recall of desired sequences (e.g. direct quotations) However, the authors note that this approach slightly decreases utility on math, knowledge and reasoning benchmarks.

Machine unlearning and ParaPO constitute post-training interventions for reducing memorization. ParaPO results in slight degradation of utility, raising the question of what extent memorization is required for utility and generalization. A future research direction may be to what extent this tradeoff can be tuned via incorporation of ``memorization" reward models that are optimized in conjunction with utility-oriented reward models, leveraging RLHF or RLVR.

\alexAA{\textbf{Model alignment} processes like instruction tuning and RLHF are primarily designed to make models safer and more helpful, but they have a significant consequential effect on memorization. As the work by Nasr et al.~\cite{nasr2023scalableextractiontrainingdata} details, these alignment procedures reduce the likelihood that a model will directly reproduce memorized training data when queried with benign or prefix-style prompts. Thus, alignment makes memorized content harder to extract under normal interaction settings. However, it does not eliminate memorization altogether: with divergence-based prompting or targeted fine-tuning, aligned models can still be induced to reproduce sensitive or verbatim sequences from pretraining data. This finding highlights that alignment primarily limits the accessibility of memorized data rather than its existence, implying that effective mitigation must combine alignment with broader privacy-preserving training or post-hoc filtering strategies. Therefore, alignment makes memorization harder to extract but does not make it harder to memorize (or prevent the memory from existing). This is a crucial distinction between achieving safety/utility and achieving true privacy/unlearning.}

\subsection{Inference-Time Interventions}

\textbf{MemFree decoding} is a strategy introduced by~\citet{ippolito-etal-2023-preventing} to filter out memorized sequences during generation. This approach contrasts with decoding by introducing a privacy filter into the generation loop applied post hoc during generation and acts as a wrapper around any pre-trained model. 
\alexAA{Ippoloto, et al.}~\citet{ippolito-etal-2023-preventing} leveraged a bloom filter to represent the set of all n-grams in the training set to identify memorized content in real time. However, near-identical n-grams are undetectable as minor modifications to a token sequence evade detection. Another limitation is that access to the LLM's n-gram training data is required.

\alexAA{TokenSwap~\cite{prashanttokenswap} introduces a lightweight, token-level method to disrupt memorized sequences at inference time. It operates entirely without access to the training corpus or model weights, addressing a key practical limitation of earlier techniques. By introducing this token-level intervention, the method effectively reduces both exact and near-verbatim generation while demonstrating easy integration with standard frameworks like HuggingFace, making it a highly scalable solution for production-level LLMs. This work highlights an emerging emphasis on practical, plug-and-play privacy interventions that can complement more computationally intensive training-time methods.}

\textbf{Activation steering} addresses memorization in LLMs by manipulating internal activations during inference~\citep{turner2023activation}. By injecting targeted perturbations into an LLM’s hidden states, activation steering can suppress or redirect the generation of memorized sequences while preserving overall model reasoning capabilities~\citep{suri2025mitigating}. \alexAA{Suri, et al.}~\citet{suri2025mitigating} demonstrated that using a sparse autoencoder to identify activation patterns linked to memorized passages enables the construction of steering directions that reduced memorization by up to 60\%, with minimal performance degradation. However, they note that steering effectiveness depends heavily on precise layer selection and steering strength.

Understanding where memorization arises in LLMs is crucial to designing effective steering interventions. \alexAA{Stoehr, et al.}~\citet{stoehr2024localizing} conducted a detailed mechanistic analysis of paragraph-length memorized sequences and found that memorization is often localized to specific LLM components. They showed that fine-tuning only a subset of the high-gradient weights was sufficient to erase memorized passages. Additionally, they identified a single attention head in an early layer that reliably activates in response to rare token combinations seen during training, triggering verbatim recall.

Building on localization, \alexAA{Chang, et al.}~\citet{chang2023localization} evaluated multiple methods for localizing memorized content by introducing two diagnostic benchmarks: injection, where known memorized sequences are inserted via fine-tuning a small set of weights, and deletion, which tests whether removing specific neurons erases a naturally memorized output. They compared several techniques, including activation-based heuristics~\citep{geva2022transformer}, integrated gradients~\citep{dai2021knowledge}, and pruning-based methods like Slimming and Hard Concrete~\citep{chang2023localization}. Their results showed that pruning-based approaches were most effective: for example, Hard Concrete was able to identify fewer than 0.5\% of neurons whose removal led to a ~60\% drop in memorization accuracy. Both \alexAA{Stoehr, et al.}~\citet{ stoehr2024localizing} and \alexAA{Chang, et al.}~\citet{chang2023localization} highlighted a key challenge: the neurons involved in one memory often contribute to others, complicating memory-specific interventions due to the risk of collateral forgetting.

By leveraging insights from model localization and sparse representation learning, such techniques can target memorized content with increasing precision, offering a flexible and minimally invasive tool for minimizing memorization.

In Appendix's Section 2.2, 
Table~\ref{table-mitigate-challenges} provides a taxonomy of challenges in memorization mitigation by distinguishing between foundational scientific problems and the practical difficulties that arise at specific stages of the model lifecycle. The \textbf{Foundational} challenges, such as disentangling knowledge from regurgitation and verifying erasure, represent the core, cross-cutting scientific obstacles. The subsequent categories, \textbf{Training-Time}, \textbf{Post-Training}, and \textbf{Inference-Time}, detail how these fundamental issues manifest as concrete engineering trade-offs, such as the utility cost of differential privacy or the risk of catastrophic forgetting during unlearning. This structure offers a comprehensive overview, connecting the deep-rooted scientific difficulties to their practical consequences at each phase of model development and deployment.

While data cleaning should be a fundamental part of every LLM training pipeline to reduce memorization, other mitigation strategies either face practicality and scalability trade-offs or lack formal guarantees of effectiveness. Memorization mitigation is typically approached from two perspectives: (1) addressing all training data, and (2) targeting specific subsets of information intended to be forgotten. Given that LLMs often indiscriminately memorize data, mitigation efforts should prioritize reducing the retention of sensitive or harmful content without compromising knowledge essential for factual recall and reasoning. From the first perspective, ensuring robust privacy protection requires evaluating the viability of scaling up differentially private (DP) training and understanding the optimal trade-off between utility and cost. Additionally, as PEFT methods grow in popularity, there is a clear need to refine DP strategies to suit these approaches and mitigate some of the overhead associated with full-model DP training. For targeted mitigation, activation steering shows promise in localizing specific knowledge representations. However, further work is needed to formalize layer selection and parameter tuning to improve our understanding and control of memorization in LLMs.

\subsection{Connections to Other Undesirable Behaviors}
It remains open whether mitigating memorization also mitigates other undesirable behaviors of LLMs, such as hallucination, racial bias, and toxic content. For instance, does reducing memorization reduce hallucination on concepts related to memorized data? Is it possible that reducing the memorization of datapoints that include harmful racial stereotypes reduces the rates at which large models produce biased or toxic content? The \alexAA{researchers} to date have not seen cross-cutting analysis examining whether reducing memorization can reduce other harmful behaviors in-tandem.

\alexAA{Mitigation efforts framed solely as the minimization of memorization overlook that an LLM's utility is dependent on its capacity to store a vast repository of factual and linguistic patterns as parametric knowledge. The challenge is not suppression but disentanglement: separating the mechanisms that support beneficial, generalizable factual recall from those that cause the harmful regurgitation of instance-specific, non-generalizable data points. This perspective re-frames the objective from one of simple elimination to one of principled, selective control, motivating the critical open questions that follow on how to define, measure, and preserve this beneficial memorization while curbing the harmful.}

\begin{tcolorbox}[
    colback=lightpurple,
    colframe=black,
    title=\textbf{Open Questions},
    fonttitle=\bfseries,
    coltitle=white,
    colbacktitle=black,
    enhanced,
    sharp corners=south,
    boxrule=0.8pt
]
1. How can we scale up DP training, and at what cost and utility trade-off? \\
2. How can DP improve PEFT to reduce memorization? \\
3. How can we optimize activation steering for memorization removal? \\
4. How can post-training methods reduce memorization while preserving utility? \\
5. Can memorization mitigation methods be incorporated online during post-training, eg. by using memorization-detecting reward models during RLHF? \\
6. When does memorization contribute to versus harm utility or generalization? \\
7. Does mitigating memorization affect other undesirable behaviors, such as hallucination or generating toxic stereotypes?
\end{tcolorbox}

\textbf{OQ1.} One direction for DP training is to develop content-sensitive DP, where the privacy budget ($\epsilon$) is allocated dynamically based on the semantic content. This reframes the challenge from applying uniform noise to intelligently learning to apply stricter protection for sensitive data while preserving utility on public knowledge, thus creating a more optimized and practical privacy-utility trade-off \cite{li2021large, aghabagherloo2025impact, SHSP2024}.

\textbf{OQ2.}
While applying DP to PEFT methods is a practical first step, a formal theory for the composability of privacy guarantees when multiple private adapters are combined is critically lacking. Future work must therefore move beyond post-hoc DP application to designing novel PEFT architectures that are private-by-construction, ensuring mathematically provable guarantees for a secure, modular model ecosystem \cite{chua2024mind, ma2024efficient, liu2024differentially}.

\textbf{OQ3.} Current activation steering methods often focus on suppressing specific, known instances of memorization. A key open question is how to scale this from single-instance intervention to learning generalizable privacy policies at the activation level. Instead of merely blocking one secret, research should focus on identifying and modifying the underlying ``reasoning paths" that lead to privacy violations, effectively teaching the model abstract rules like ``do not reveal PII" that generalize  \cite{meng2022locating, suri2025mitigating, turner2023activation, stoehr2024localizing}.

\textbf{OQ4.} As the machine unlearning paradigm shifts from one-shot erasures to a continuous governance framework, it creates critical challenges in developing scalable and formally verifiable deletion methods. Future work must therefore produce efficient and auditable techniques designed for the ongoing maintenance of LLMs as dynamic systems, capable of processing numerous requests without catastrophic forgetting \cite{nguyen2025survey, cooper2024machine, chen2025parapoaligninglanguagemodels}.

\textbf{OQ5.} The next evolution beyond a static memorization-detecting reward model is to create a dynamic adversarial alignment process. This would involve co-training a ``red team" AI designed to continuously find and exploit memorization vulnerabilities, while the main LLM is rewarded for evading these adaptive attacks. This moves beyond fixed-preference optimization to a more robust, game-theoretic approach to achieving privacy during alignment \cite{schulman2017proximal, rafailov2024directpreferenceoptimizationlanguage}.

\textbf{OQ6.} The binary ``good" versus ``bad" memorization is insufficient; the future requires quantifying a spectrum of generalizability for all knowledge within an LLM. Research should develop information-theoretic or causal metrics to place any model output on a continuum from a non-generalizable, instance-specific regurgitation to a broadly applicable, foundational fact. This reframes the problem from a simple classification to a nuanced, risk-based assessment of knowledge provenance \cite{feldman2020neural, hong2025reasoning, zhang2023counterfactual}.

\textbf{OQ7.} Instead of tackling memorization, hallucination, and toxicity as separate problems, research should investigate a unified mechanistic hypothesis: that they are all symptoms of over-relying on spurious statistical correlations. Developing training methods rooted in causal representation learning could be a single, powerful intervention that simultaneously reduces harmful memorization, improves factuality, and curtails toxic stereotypes by forcing the model to learn the true data-generating process \cite{bender2021dangers, siau2020artificial}.

\section{Privacy \& legal risks of memorization}
Memorization in LLMs poses serious security and privacy risks: memorized text can be directly leaked, undermining data confidentiality~\citep{smith2023identifying}. We discuss three major impact areas caused by memorization: personal data leakage, exposure of copyrighted or proprietary content, and broader legal and public consequences.
%

\textbf{Personal data leakage.} As explored in~\citet{carlini2021extractingtrainingdatalarge, li2024llmpbeassessingdataprivacy, jagielski2023measuring, zhou2024quantifying}, a significant risk for LLMs is leaking sensitive personal information.
In~\citet{panda2024teach}, researchers demonstrated a targeted neural phishing attack that achieved up to a 50\% success rate in tricking a model into revealing PII during fine-tuning when injecting poisons during pretraining. Even when training data is intended to remain private, malicious attackers with the correct model access level can siphon memorized secrets~\citep{panda2024teach}.
If an LLM reveals any PII, this could result in identity theft or other harms. Furthermore, for industries such as healthcare or customer service, where customers expect confidentiality, this would create significant liability for model operators under current privacy regulations. 

\textbf{Copyright or proprietary content.} Beyond PII, memorization in LLMs may also cause the reproduction of copyrighted or proprietary material such as books, code, etc. This turns the LLM into a potential conduit for the unauthorized distribution of protected content. As observed by~\citet{carlini2023quantifyingmemorizationneurallanguage}, models may produce verbatim memorization from their training data. While we can only explicitly confirm what training data is used for open-source LLMs~\citep{gao2020pile, soldaini2024dolma}, there do exist copyrighted books in EleutherAI’s PILE dataset, specifically, copyrighted books in its Books3 section. This is observed in~\citet{henderson2023foundation, zhang2022opt}, where open models can regurgitate the first few pages of the Harry Potter books and ``Oh, the Places You'll Go!" by Dr. Seuss verbatim, raising concerns.

\textbf{Legal landscape.} There are several ongoing cases that will shape the legal landscape of LLMs. The New York Times v. Microsoft Corp. lawsuit~\citep{nytimes_v_microsoft_2023} represents a pivotal legal challenge to generative AI and copyright, accompanied by similar cases, including Chabon v. OpenAI~\citep{chabon_v_openai_2023} where authors claiming copyright infringement in model training, and Doe v. GitHub~\citep{doe_v_github_2022} where plaintiffs allege verbatim code reproduction in violation of the Digital Millennium Copyright Act (DMCA). The Times claimed that OpenAI’s model trained on their copyrighted articles without proper license and that ChatGPT could regurgitate passages of those articles on request. In~\citet{nytimes_v_microsoft_2023}, the publishers argue this behavior amounts to mass copyright infringement since the model’s outputs mimic and even compete with the Times’ content. High-profile copyright litigation, exemplified by~\citep{nytimes_v_microsoft_2023}, has thrust LLM memorization from a technical concern into the public policy and business arenas, a shift underscored by empirical analyses of the lawsuit's claims from privacy researchers~\citep{freeman2024exploringmemorizationcopyrightviolation}.

In Appendix's Section 2.3, Table~\ref{table-legal-risk-challenges} provides a taxonomy that traces the path from core technical failures to their downstream legal and societal risks. It illustrates how specific \textbf{Technical Challenges}, such as the inability to guarantee verifiable erasure, directly create \textbf{Legal \& Regulatory Challenges}, like the difficulty of complying with General Data Protection Regulation (GDPR)'s Right to Erasure, which in turn manifest as tangible \textbf{Socio-Technical Challenges}, such as the erosion of public trust. This structure is applied consistently across all impact areas, from personal data leakage to copyright infringement. The table's central insight is that legal and societal risks are not independent but are fundamentally constrained by underlying technical limitations.

\begin{tcolorbox}[
    colback=lightpurple,
    colframe=black,
    title=\textbf{Open Questions},
    fonttitle=\bfseries,
    coltitle=white,
    colbacktitle=black,
    enhanced,
    sharp corners=south,
    boxrule=0.8pt
]
1. Can LLMs be trained to avoid memorizing copyrighted content? \\
2. How can memorization metrics inform legal analysis (fair use, copyright)? \\
3. What technical threshold defines legally significant memorization?
\end{tcolorbox}

\textbf{OQ1.} The next research frontier is to move beyond simple data filtering and train LLMs to be copyright-aware. This involves augmenting training corpora with copyright metadata and modifying the optimization objective to explicitly penalize the generation of text that is both verbatim and tagged as protected. This approach reframes copyright compliance from a static data-curation problem into a learnable, dynamic behavior, enabling the model to respect intellectual property as an intrinsic part of its generation process \cite{vyas2023provable, karamolegkou2023copyright}.

\textbf{OQ2.} Future memorization metrics must evolve from measuring superficial string similarity to quantifying the degree of transformative use, a cornerstone of fair use doctrine. The research challenge is to develop computational methods that assess not just what was reproduced, but how it was used—for example, distinguishing a derivative summary (potentially fair use) from a substitutive replication (potential infringement). This requires creating metrics that capture the functional and semantic relationship between an LLM's output and a copyrighted source, providing a technical basis for legal arguments \cite{henderson2023foundation, schwarzschild2024rethinking}.

\textbf{OQ3.} Instead of searching for a single, universal threshold, the research community should develop a framework for context-aware memorization auditing. This framework would replace a simple percentage with a multi-faceted report assessing factors like the novelty of the memorized text content, its commercial value, and the ``extraction effort" required by a user to elicit it. This re-orients the goal from finding a magic number to providing a rich, evidence-based dossier that legal experts can use to argue ``substantial similarity" on a case-by-case basis \cite{cooper2023report, nytimes_v_microsoft_2023}.

\section{Conclusion}
Memorization in LLMs represents a double-edged sword in the advancement of AI systems. This phenomenon, best understood not as a bug but as a misaligned feature of data compression, creates a fundamental tension between model utility and privacy/legal concerns that must be addressed.

This perspective clarifies the limitations of current methods and illuminates the path forward. Detection techniques must move beyond identifying symptoms to enable true attribution of knowledge. Mitigation strategies like de-duplication and unlearning, which aim for erasure, show potential but are insufficient; the frontier lies in developing robust methods for control over what a model recalls. Ongoing legal challenges highlight the need for frameworks that can manage this control and attribution to preserve beneficial knowledge while preventing information leakage.

By advancing our understanding in these areas, we can work toward building more powerful and trustworthy AI systems that maintain high performance while respecting privacy and intellectual property concerns. This new focus on principled control and attribution, rather than a simple balance, will be essential as LLMs continue to be integrated into increasingly sensitive applications.

\bibliography{main}
\bibliographystyle{IEEEtran}

\newpage
\appendix

\subsection{Definitions of memorization}
\label{def-memorization-cont}
\subsubsection{Exact memorization}

\textbf{Verbatim memorization} refers to a model's exact reproduction of training data, often arising due to the high duplication of specific examples or overfitting~\citep{carlini2021extractingtrainingdatalarge}. 
This phenomenon occurs when an LLM, instead of generating novel responses or synthesizing information, simply retrieves and reproduces specific text snippets word-for-word from the data it was exposed to.

\textbf{Perfect memorization} 
describes a model that has exactly recorded its training data, assigning a probability only to inputs it has previously seen~\citep{kandpal2022deduplicatingtrainingdatamitigates}. When such a model generates an output, the process is mathematically equivalent to randomly selecting examples from its training dataset with a uniform probability. 

\textbf{Eidetic memorization}~\citep{carlini2021extractingtrainingdatalarge} occurs when a prompt $p$ causes the model to reproduce a verbatim string $s$ from training data. A stricter variant, \textbf{$k$-eidetic memorization}~\citep{carlini2021extractingtrainingdatalarge}, occurs when $s$ appears in no more than $k$ training examples, yet can still be extracted given the right prompt.

\textbf{Discoverable memorization}. A string $s$ is discoverably memorized~\citep{carlini2023quantifyingmemorizationneurallanguage,nasr2023scalableextractiontrainingdata} 
when, given a training example composed of prefix $p$ and suffix $s$, an LLM exactly reproduces the continuation $s$ that followed $p$ in the training data.

\subsubsection{Approximate memorization}
\textbf{Approximate/Paraphrased memorization}~\citep{ippolito-etal-2023-preventing} 
is when LLMs generate outputs similar to training data in content, structure, or phrasing without exact replication. This differs from exact copying through variations, paraphrasing, or partial overlaps. 
The authors detect memorization by calculating edit distance between a generation and a target string, normalized by length, and choosing a threshold that determines when a sequence is deemed an approximate match.

\subsubsection{Prompt-based memorization}
\textbf{Extractable memorization} 
occurs when, without access to the training data, there exists a constructable prompt that invokes the model to generate an exact example from its training set~\citep{nasr2023scalableextractiontrainingdata}. 

\textbf{k-extractable memorization} represents a stricter form of extractable memorization. A suffix is k-extractable~\citep{biderman2023emergentpredictablememorizationlarge} 
when, given only its corresponding k-token prefix, the model reproduces the entire suffix verbatim. Unlike verbatim memorization, which manifests as direct replication, k-extractable memorization captures a model's ability to retrieve and complete specific training sequences when prompted with only partial context. 

\textbf{($n, p$)-discoverable extraction}~\citep{hayes2025measuringmemorizationlanguagemodels} formalizes the likelihood of retrieving a memorized string via repeated sampling. A string is ($n, p$)-discoverable if it appears in at least one of $n$ completions with probability  $\geq p$.

\subsubsection{Influence-based memorization}
\textbf{Counterfactual memorization}~\citep{feldman2020neural, zhang2023counterfactual} 
quantifies how much a model's predictions change based on whether a model is trained on a particular training example. For \alexAA{ instance}, this can be instantiated as how much the loss changes on a datapoint when seen by a model during training versus not. \alexAA{Pappu, et al.}~\citet{pappu2024measuringmemorizationrlhfcode} instantiate this definition by classifying datapoints as memorized when both a model trained on the datapoint achieves an approximate match under prefix-based decoding and a model not trained on that datapoint does not achieve an approximate match. As it is computationally prohibitive to retrain a model per datapoint-exlusion,  \alexAA{Pappu, et al.}~\citet{pappu2024measuringmemorizationrlhfcode} and \alexAA{Zhang, et al.}~\citet{zhang2023counterfactual} train a smaller number of models that exclude entire subsets of data used to test for memorization.


\subsection{Tables for Research Challenges in Detection, Mitigation, and Privacy and Legal Risks}

\subsubsection{Table~\ref{table-detect-challenges}}
\label{table-appendix-detect-challenge}
It outlines the primary research challenges for
detecting memorization.

\begin{table*}[h!]
\footnotesize
\centering
\resizebox{\linewidth}{!}{
\begin{tabular}{ p{4.5cm}|p{5.5cm}|p{6.5cm} } 
\toprule
{\bf Category} & {\bf Research Challenges} & {\bf Impact} \\
\midrule
Defining the Detection Target
&
\begin{itemize}[leftmargin=*]\vspace{-2mm}
\item Disentangling parametric vs. retrieved knowledge~\citep{bai2024special}. \item 
Formalizing memorization equivalence~\citep{schwarzschild2024rethinking}.
\end{itemize}
&
\begin{itemize}[leftmargin=*]\vspace{-2mm}
    \item Creates  ambiguity in the detection target (parametric vs. retrieved, verbatim vs. paraphrased).
    \item Threatens the validity of detectors for real-world tasks like copyright and privacy auditing~\citep{ippolito-etal-2023-preventing}.
\end{itemize}
\\
\midrule
Methodological Rigor \& Reliability
&
\begin{itemize}[nosep, leftmargin=*]\vspace{-2mm}
\item 
Calibrating MIA~\citep{zhang2025membershipinferenceattacksprove} detection thresholds across models and domains~\citep{chen2024statistical}. 
\item Lack of standardized comparative benchmarks.
\end{itemize}
&
\begin{itemize}[leftmargin=*]\vspace{-2mm}
    \item Hinders rigorous verification by preventing meaningful, standardized comparisons between detection techniques.
\end{itemize}
\\
\midrule
Scope \& Viability
&
\begin{itemize}[leftmargin=*]\vspace{-2mm}
\item Prohibitive computational overhead of rigorous methods (e.g., shadow models). 
\item Low precision on high-stakes tasks like character-level PII extraction~\citep{nakka2024pii}.
\end{itemize}
&
\begin{itemize}[leftmargin=*]\vspace{-2mm}
    \item Large-scale auditing infeasible for most organizations due to prohibitive computational costs.
\end{itemize}
\\
\bottomrule
\end{tabular}
}\vspace{3mm}
\caption{A Taxonomy of Research Challenges in Memorization Detection.}
\label{table-detect-challenges}
\end{table*}

\subsubsection{Table~\ref{table-mitigate-challenges}}
\label{table-appendix-mitigation-challenge}

It presents the primary research challenges for mitigating memorization.

\begin{table*}[h!]
\footnotesize
\centering
\setlist[itemize]{nosep, leftmargin=*, topsep=0pt, partopsep=0pt}
\resizebox{\linewidth}{!}{
\begin{tabular}{p{2.5cm}|p{4.5cm}|p{11.5cm}} 
\toprule
{\bf Category} & {\bf Key Research Challenges} & {\bf Problem Specifications \& Impact}  \\
\midrule
\multirow{4}{*}{\textbf{Foundational}} &
\textbf{Disentangling Knowledge from Regurgitation}~\citep{hong2025reasoning}
&
\begin{minipage}[t]{\linewidth}
\begin{itemize}
    \item Mechanisms for useful factual recall are entangled with those for harmful data regurgitation~\citep{huang2024demystifying}.
    \item Lack of architectural separation between general semantic knowledge and instance-specific, episodic data.
\end{itemize}
\end{minipage}
\\
\cmidrule{2-3}
&
\textbf{Verifiability and Proof of Forgetting}~\citep{xue2025towards}
&
\begin{minipage}[t]{\linewidth}
\begin{itemize}
    \item Absence of formal guarantees to certify that a data point's influence has been completely erased~\citep{eisenhofer2025verifiable}.
    \item Makes unlearning methods difficult to audit and trust, especially for regulatory compliance and against adversarial verification~\citep{zhang2024verification}.
\end{itemize}
\end{minipage}
\\
\midrule
\multirow{4}{*}{\textbf{Training-Time}} &
\textbf{Scalability and Utility of Differential Privacy}~\citep{li2021large}
&
\begin{minipage}[t]{\linewidth}
\begin{itemize}
    \item Prohibitive computational cost of applying techniques like DP-SGD to web-scale training corpora~\citep{abadi2016deep}.
    \item Stronger privacy guarantees often lead to significant degradation in model performance and utility.
\end{itemize}
\end{minipage}
\\
\cmidrule{2-3}
&
\textbf{Semantic Depth of Data Sanitization}~\citep{SHSP2024}
&
\begin{minipage}[t]{\linewidth}
\begin{itemize}
    \item Surface-level de-duplication (e.g., hash-based) is insufficient for privacy, as models still memorize near-duplicates~\citep{lee2022deduplicatingtrainingdatamakes}.
    \item Models remain vulnerable to memorizing paraphrased or semantically similar sensitive information that bypasses simple filters~\citep{ippolito-etal-2023-preventing}.
\end{itemize}
\end{minipage}
\\
\midrule
\multirow{4}{*}{\textbf{Post-Training}} &
\textbf{Preserving Utility During Unlearning}~\citep{huang2024demystifying}
&
\begin{minipage}[t]{\linewidth}
\begin{itemize}
    \item Unlearning one piece of information can degrade unrelated model capabilities~\citep{kassem2023preserving}.
    \item Can lead to catastrophic forgetting, where the model loses significant general knowledge that was not targeted for removal~\citep{anjarlekar2025llm}.
\end{itemize}
\end{minipage}
\\
\cmidrule{2-3}
&
\textbf{Durability Against Relearning Attacks}~\citep{fan2025towards}
&
\begin{minipage}[t]{\linewidth}
\begin{itemize}
    \item The ``unlearned" state of a model is often fragile and not permanent~\citep{cooper2024machine}.
    \item Small exposures to previously forgotten data can quickly reverse the unlearning, undermining long-term effectiveness~\citep{fan2025towards}.
\end{itemize}
\end{minipage}
\\
\midrule
\multirow{4}{*}{\textbf{Inference-Time}} &
\textbf{Generalizability of Activation Steering}~\citep{suri2025mitigating}
&
\begin{minipage}[t]{\linewidth}
\begin{itemize}
    \item Interventions are often brittle and highly tailored to specific data instances or activation patterns~\citep{chang2023localization}.
    \item Fails to prevent leakage of semantically equivalent information or other, untargeted secrets.
\end{itemize}
\end{minipage}
\\
\cmidrule{2-3}
&
\textbf{Efficiency and Coverage of Secure Decoding}~\citep{prashanttokenswap}
&
\begin{minipage}[t]{\linewidth}
\begin{itemize}
    \item Introduces significant computational latency at inference time, making it impractical for many real-time applications.
    \item Easily circumvented by minor paraphrasing or stylistic changes to the output, providing a false sense of security~\citep{ippolito-etal-2023-preventing}.
\end{itemize}
\end{minipage}
\\
\midrule
\textbf{Cross-Cutting} &
\textbf{Extending Mitigation to Multimodal Models}~\citep{wen2025quantifyingcrossmodalitymemorizationvisionlanguage}
&
\begin{minipage}[t]{\linewidth}
\begin{itemize}
    \item Existing mitigation techniques are overwhelmingly designed for text-only models.
    \item Lack of clear definitions and methods for handling cross-modal memorization (e.g., text prompt eliciting a memorized image)~\citep{qii2025}.
\end{itemize}
\end{minipage}
\\
\bottomrule
\end{tabular}
}\vspace{3mm}
\caption{A Taxonomy of Challenges in Memorization Mitigation.}
\label{table-mitigate-challenges}
\end{table*}

\subsubsection{Table~\ref{table-legal-risk-challenges}}
\label{tables-privacy-risks}

It summarizes the primary research challenges for memorization at the intersection of privacy, regulatory risks, and societal impact.

\begin{table*}[h!]
\footnotesize
\centering
\resizebox{\linewidth}{!}{
\begin{tabular}{ p{2.8cm} | p{4.5cm} | p{4.5cm} | p{4.5cm} } 
\toprule
 & \multicolumn{3}{c}{\bf Taxonomy of Challenges in Privacy and Legal Risks} \\
\cmidrule{2-4}
\textbf{Impact Area} & \textbf{Technical Challenges} & \textbf{Legal \& Regulatory Challenges} & \textbf{Socio-Technical Challenges} \\
\midrule
\textbf{Personal Data Leakage} &
\begin{itemize}[nosep, leftmargin=*]
    \item \textbf{Inseparability:} Sensitive data is deeply entangled in model parameters, making precise isolation and removal technically infeasible~\citep{nakka2024pii}.
    \item \textbf{Conceptual Unlearning Failure:} Token-level unlearning is easily bypassed by rephrased prompts, failing to remove the underlying concept~\citep{slonski2024detecting}.
    \item \textbf{Lack of Verifiable Erasure:} There are no provable guarantees that empirical unlearning has completely and permanently removed data influence~\citep{kassem2023preserving}.
\end{itemize}
&
\begin{itemize}[nosep, leftmargin=*]
    \item \textbf{Right to Erasure (GDPR):} Fulfilling data deletion requests is technically challenging due to the opaque and distributed nature of LLM parameters.
    \item \textbf{Evolving Definition of PII:} The scope of what constitutes personal data that a model can infer (not just regurgitate) is legally ambiguous in the US and is inconsistent with GDPR definitions.
\end{itemize}
&
\begin{itemize}[nosep, leftmargin=*]
    \item \textbf{Erosion of User Trust:} The inability to guarantee data deletion undermines user confidence and creates significant reputational risk for model deployers.
    \item \textbf{High Mitigation Cost vs. Utility:} The trade-off between privacy, model performance, and computational cost makes effective protection economically prohibitive.
\end{itemize}
\\
\midrule
\textbf{Copyright \& Proprietary Content} &
\begin{itemize}[nosep, leftmargin=*]
    \item \textbf{Paraphrastic Leakage:} Detection methods are inadequate for identifying paraphrased or stylistically similar reproductions, which are key to copyright disputes~\citep{schwarzschild2024rethinking}.
    \item \textbf{Prohibitive Mitigation Cost:} Retraining models from scratch to remove infringing content is economically and computationally infeasible~\citep{anjarlekar2025llm}.
    \item \textbf{Fragility of Forgetting:} Unlearned models can rapidly re-memorize copyrighted works from minimal exposure, undermining long-term compliance~\citep{dong2025practical}.
\end{itemize}
&
\begin{itemize}[nosep, leftmargin=*]
    \item \textbf{Legal Ambiguity of ``Fair Use":} The line between transformative use and derivative infringement for generative models is legally unsettled.
    \item \textbf{Liability for Infringement:} Ongoing high-profile lawsuits are shaping the legal landscape, creating uncertainty for developers and deployers.
\end{itemize}
&
\begin{itemize}[nosep, leftmargin=*]
    \item \textbf{Degradation of Model Quality:} Suppressing verbatim outputs can inadvertently harm a model's creative and factual capabilities, making it less useful~\citep{huang-etal-2024-demystifying}.
\end{itemize}
\\
\midrule
\textbf{Broader Legal \& Public Consequences} &
\begin{itemize}[nosep, leftmargin=*]
    \item \textbf{Reasoning vs. Regurgitation:} It is technically difficult to determine if a model is genuinely reasoning or just ``stitching together" memorized fragments, a key issue for reliability~\citep{hong2025reasoning}.
    \item \textbf{Functional Knowledge Unlearning:} A major gap exists in methods to unlearn a ``skill" (e.g., using a deprecated API) versus a specific data point~\citep{suri2025mitigating, huang2025df}.
\end{itemize}
&
\begin{itemize}[nosep, leftmargin=*]
    \item \textbf{Lack of Technical Standards for Law:} The legal system has no established technical threshold for what constitutes legally significant memorization.
    \item \textbf{Compliance with Future Regulation:} Models must be designed to adapt to a rapidly evolving global regulatory landscape for AI.
\end{itemize}
&
\begin{itemize}[nosep, leftmargin=*]
    \item \textbf{Breakdown of Public Trustworthiness:} If users cannot trust that a model's output is the result of genuine reasoning, its utility in high-stakes domains (e.g., medicine, law) may be critically impaired~\citep{suri2025mitigating,li2025memorization}.
    \item \textbf{Accountability Gap:} The opaque nature of memorization makes it difficult to assign responsibility when a model produces harmful or illegal content.
\end{itemize}
\\
\bottomrule
\end{tabular}
}\vspace{3mm}
\caption{A Taxonomy of Challenges at the Intersection of Memorization, Privacy, and Legal Risks.}
\label{table-legal-risk-challenges}
\end{table*}

\end{document}